\documentclass[sigconf]{acmart}

\usepackage{booktabs} 
\usepackage{subfigure}
\usepackage{footnote}
\makesavenoteenv{tabular}
\makesavenoteenv{table}
\setcopyright{none}
\renewcommand\footnotetextcopyrightpermission[1]{}
\acmConference[MLMH Workshop, KDD]{Machine Learning for Medicine and Healthcare Workshop}{August 19-23, 2018}{London}
\acmYear{2018}

\begin{document}
\title{Transfer Learning for Clinical Time Series Analysis using Recurrent Neural Networks}

\author{Priyanka Gupta,
Pankaj Malhotra, 
Lovekesh Vig, 
Gautam Shroff}
\affiliation{\institution{TCS Research, New Delhi, India}}
\email{{priyanka.g35, malhotra.pankaj, lovekesh.vig, gautam.shroff}@tcs.com}

\renewcommand{\shortauthors}{Gupta, Malhotra, Vig, Shroff}
\renewcommand{\shorttitle}{Transfer Learning for Clinical Time Series using RNNs}

\begin{abstract}
Deep neural networks have shown promising results for various clinical prediction tasks such as diagnosis, mortality prediction, predicting duration of stay in hospital, etc. 
However, training deep networks -- such as those based on Recurrent Neural Networks (RNNs) -- requires large labeled data, high computational resources, and significant hyperparameter tuning effort.
In this work, we investigate as to what extent can transfer learning address these issues when using deep RNNs to model multivariate clinical time series.
We consider transferring the knowledge captured in an RNN trained on several source tasks simultaneously using a large labeled dataset to build the model for a target task with limited labeled data.
An RNN pre-trained on several tasks provides generic features, which are then used to build simpler linear models for new target tasks without training task-specific RNNs.
For evaluation, we train a deep RNN to identify several patient phenotypes on time series from MIMIC-III database, and then use the features extracted using that RNN to build classifiers for identifying previously unseen phenotypes, and also for a seemingly unrelated task of in-hospital mortality.
We demonstrate that (i) models trained on features extracted using pre-trained RNN outperform or, in the worst case, perform as well as task-specific RNNs; 
(ii) the models using features from pre-trained models are more robust to the size of labeled data than task-specific RNNs; and
(iii) features extracted using pre-trained RNN are generic enough and perform better than typical statistical hand-crafted features.
\end{abstract}


\maketitle

\section{Introduction}
Electronic health records (EHR) consisting of a patient's medical history can be leveraged for various clinical applications such as diagnosis, recommending medicine, etc. Traditional machine learning techniques often require careful domain-specific feature engineering before building the prediction models.
On the other hand, deep learning approaches enable end-to-end learning without the need of hand-crafted and domain-specific features, and have recently produced promising results for various clinical prediction tasks \cite{miotto2017deep,ravi2017deep}.
Applications of such approaches include medical diagnosis \cite{choi2016doctor,lipton2015learning}, predicting future clinical events \cite{miotto2016deep}, etc.

Deep Recurrent Neural Networks (RNNs) have been successfully explored for various time series and sequential modeling applications of EHR data such as diagnoses \cite{lipton2015learning,che2016recurrent,choi2016doctor}, mortality prediction and estimating length of stay \cite{harutyunyan2017multitask,purushotham2017benchmark,rajkomar2018scalable}.
However, training RNNs is compute-intensive due to sequential nature of computations, and requires large amount of labeled data. 
Transfer learning \cite{pan2010survey,bengio2012deep} has been used to overcome these challenges. 
It enables knowledge transfer from neural networks trained on a \textit{source} task with sufficient training instances to a related \textit{target} task with few training instances.
Moreover, fine-tuning a pre-trained network for target task is often faster and easier than constructing and training a new network from scratch \cite{bengio2012deep,malhotra2017timenet}.
Another advantage of learning in such a manner is that the pre-trained network has already learned to extract a rich set of generic features that can then be applied to a wide range of other similar tasks \cite{malhotra2017timenet,gupta2018using}.

Transfer learning via fine-tuning parameters of pre-trained models for end tasks has been recently considered for medical applications, e.g. \cite{choi2016doctor,lee2017transfer}.
However, fine-tuning a large number of parameters with a small labeled dataset may cause overfitting. 
If the parameters to be tuned for target task can be reduced to a small number, then the pre-trained deep models can be leveraged in a better way \cite{keshari2018learning}.
In this work, \textit{we evaluate an approach to transfer the learning from a set of tasks to another related task for clinical time series by means of an RNN.}
Considering phenotype detection from time series of physiological parameters as a binary classification task, we train an RNN classifier on a diverse set of such binary classification tasks (one task per phenotype) simultaneously using a large labeled dataset; so that the RNN thus obtained provides general-purpose features for time series.
The features extracted using this RNN are then transferred to train a simple logistic regression model \cite{hosmer2013applied} for target tasks, i.e. identifying a new phenotype and predicting in-hospital mortality, with few labeled instances (detailed in Section \ref{sec:approach}).

Through empirical evaluation on MIMIC-III dataset \cite{johnson2016mimic} (as detailed in Section \ref{sec:exp}), we demonstrate that:
1) it is possible to leverage deep RNNs for clinical time series classification tasks in scarcely-labeled scenarios via transfer learning;
2) a deep model trained on multiple diverse tasks on a large labeled dataset provides features that are generic enough to build models for new tasks from clinical time series data.
Further, our approach provides a computationally-efficient way to use deep models for new phenotypes once an RNN has been trained to classify a diverse-enough set of phenotypes.

\section{Related Work}\label{sec:rw}
Unsupervised pre-training has been shown to be effective in capturing the generic patterns and distribution from EHR data \cite{miotto2016deep}. 
Further, RNNs for time series classification from EHR data have been successfully explored, e.g. in \cite{lipton2015learning,che2016recurrent}.
However, these approaches do not address the challenge posed by limited labeled data, which is the focus of this work.
Transfer learning using deep neural networks has been recently explored for medical applications: 
A model learned from one hospital could be adapted to another hospital for same task via recurrent neural networks \cite{choi2016doctor}.
A deep neural network was used to transfer knowledge from one dataset to another while the source and target tasks (named-entity recognition from medical records) are the same in \cite{lee2017transfer}.
However, in both these transfer learning approaches, the source and target tasks are the same while only the dataset changes. 

Features extracted from a pre-trained off-the-shelf RNN-based feature extractor (TimeNet \cite{malhotra2017timenet}) have been  shown to be useful for patient phenotyping and mortality prediction tasks \cite{gupta2018using}.
In this work, we provide an approach to transfer the model trained on several healthcare-specific tasks to a different (although related) classification task using RNNs for clinical time series.
Training a deep RNN for multiple related tasks simultaneously on clinical time series has been shown to improve the performance for all tasks \cite{harutyunyan2017multitask}.
In this work, we additionally demonstrate that a model trained in this manner serves as a reasonable starting point for building models for new related tasks.

\section{Proposed Approach} \label{sec:approach}
\begin{figure}[th]
\centering
\includegraphics[trim={1cm 0.4cm 10.2cm 5cm},clip,width=\columnwidth]{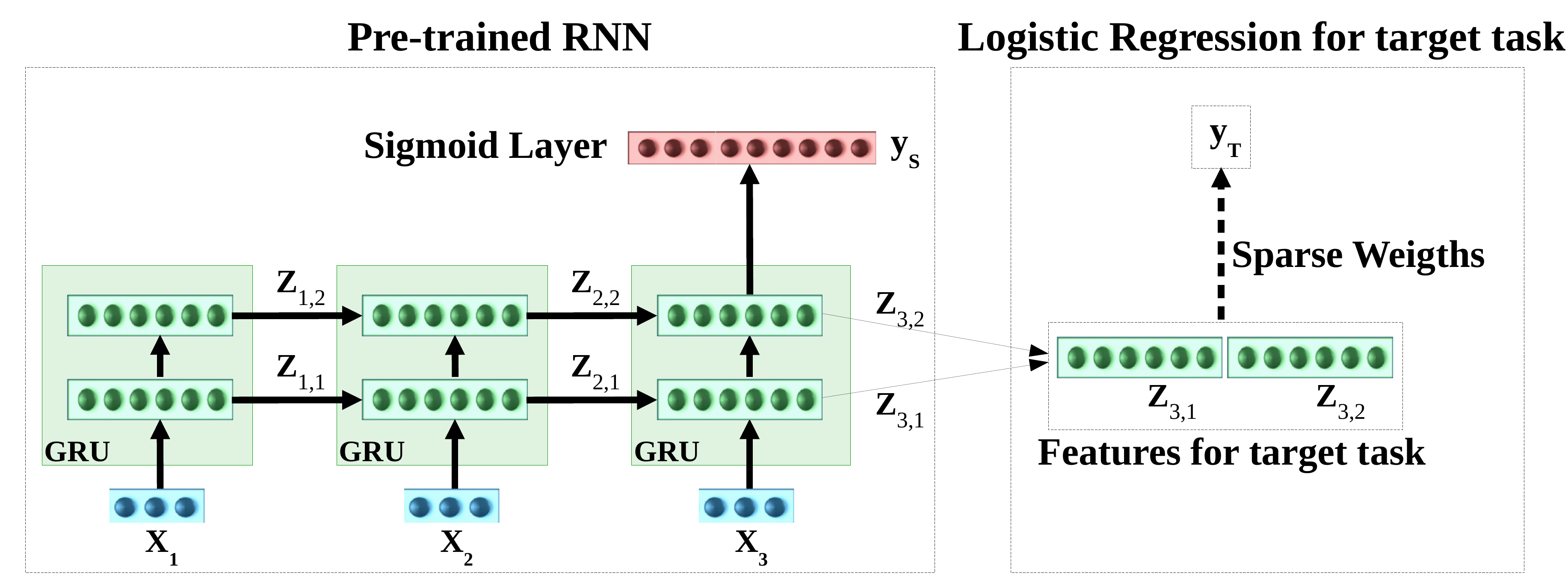}
\caption{\label{fig:multilabel} Inference in the proposed transfer learning approach. RNN with $L=2$ hidden layers is shown unrolled over $\tau=3$ time steps.}
\end{figure}

Consider sets $\mathcal{D}_S$ and $\mathcal{D}_T$ of labeled time series instances corresponding to source ($S$) and target ($T$) tasks, respectively.
$\mathcal{D}_S=\{(\mathbf{x}_S^{(i)},\mathbf{y}_S^{(i)})\}_{i=1}^{N_S}$, where $N_S$ is the number of time series instances corresponding to $N_S$ patients (in our experiments, we consider each episode of hospital stay for a patient as a separate data instance). 
Denoting time series $\mathbf{x}^{(i)}_S$ by $\mathbf{x}$ and the corresponding target label $\mathbf{y}_S^{(i)}$ by $\mathbf{y}$ for simplicity of notation, we have $\mathbf{x}=\mathbf{x}_1\mathbf{x}_2\ldots\mathbf{x}_{\tau}$ denote a time series of length $\tau$, where $\mathbf{x}_t\in \mathbb{R}^n$ is an $n$-dimensional vector corresponding to $n$ parameters such as glucose level, heart rate, etc.
Further, $\mathbf{y}=[y_1,\ldots,y_K] \in \{0,1\}^K$, where $K$ is the number of binary classification tasks. 
For example, for $K=5$ binary classification tasks corresponding to presence or absence of 5 phenotypes, $\mathbf{y}=[1,0,1,1,0]$ indicates that phenotypes 1, 3, and 4 are present while phenotypes 2 and 5 are absent.
$\mathcal{D}_T=\{(\mathbf{x}_T^{(i)},y_T^{(i)})\}_{i=1}^{N_T}$ such that $N_T \ll N_S$, and $y_T^{(i)}\in\{0,1\}$ such that the target task is a binary classification task.
We assume that the time series in $\mathcal{D}_T$ belongs to same $n$ parameters as in $\mathcal{D}_S$.
We first train the deep RNN on $K$ source tasks using $\mathcal{D}_S$, and then train the simpler logistic regression (LR) classifier for target task using $\mathcal{D}_T$ and the features obtained via the deep RNN, as shown in Figure \ref{fig:multilabel}.
We next provide details of training RNN and LR models.

\subsection{Supervised Pre-training of RNN}
Training an RNN on $K$ binary classification tasks simultaneously can be considered as a multi-label classification problem.
We train a multi-layered RNN with $L$ recurrent layers having Gated Recurrent Units (GRUs) \cite{cho2014learning} to map $\mathbf{x}^{(i)} \in \mathcal{D}_S$ to $\mathbf{y}^{(i)}$.
Let $\mathbf{z}_{t,l}\in \mathbb{R}^h$ denote the output of recurrent units in $l$-th hidden layer at time $t$, and $\mathbf{z}_{t}=[\mathbf{z}_{t,1},\ldots,\mathbf{z}_{t,L}] \in \mathbb{R}^m$ denote the hidden state at time $t$ obtained as concatenation of hidden states of all layers, where $h$ is the number of GRU units in a hidden layer and $m=h\times L$.
The parameters of the network 
are obtained by minimizing the cross-entropy loss given by $\mathcal{L}$ via stochastic gradient descent: 
\begin{equation}\label{eq:RNN-C}
\begin{aligned}
\mathbf{z}^{(i)}_\tau&=f_E(\mathbf{x}^{(i)};\mathbf{W}_E),~
\mathbf{\hat{y}}^{(i)} = \sigma(\mathbf{W}_C\:\mathbf{z}^{(i)}_{\tau,L}+\mathbf{b}_C)\\
C(y_k^{(i)},\hat{y}_k^{(i)})&=y_k^{(i)}\cdot log(\hat{y}_k^{(i)})+(1-y_k^{(i)})\cdot log((1-\hat{y}_k^{(i)}))\\
\mathcal{L}&=-\frac{1}{N_S\times K}\sum_{i=1}^{N_S}\sum_{k=1}^{K}C(y_k^{(i)},\hat{y}_k^{(i)}).
\end{aligned}
\end{equation}
Here $\sigma({x})$ =  $({1+e^{-x}})^{-1}$ is the sigmoid activation function, $\mathbf{\hat{y}}^{(i)}$ is the estimate for target $\mathbf{y}^{(i)}$, $\mathbf{W}_E$ are parameters of recurrent layers, and $\mathbf{W}_C$ and $\mathbf{b}_C$ are parameters of the classification layer.

\subsection{Using features from pre-trained RNN\label{ssec:LR}}
For input $\mathbf{x}^{(i)} \in \mathcal{D}_T$, the hidden state $\mathbf{z}^{(i)}_{\tau}$ at last time step $\tau$ is used as input feature vector for training the LR model. 
We obtain probability of the positive class for the binary classification task as $\hat{y}^{(i)}$ = $\sigma(\mathbf{w}'_C\:\mathbf{z}^{(i)}_{\tau}+{b}'_C)$, where $\mathbf{w}'_C$, ${b}'_C$ are parameters of LR.
The parameters are obtained by minimizing the negative log-likelihood loss $\mathcal{L'}$: 

\begin{equation}\label{eq:RNN-C-tune}
\begin{aligned}
\mathcal{L'}&=-\sum_{i=1}^{N_T}C(y^{(i)},\hat{y}^{(i)})+ \lambda\mathbf{\|{w}'}_C\|_1\\
\end{aligned}
\end{equation}
where $||\mathbf{w}'_C||_1=\sum_{j=1}^m|w_{j}|$ is the L$_1$ regularizer with $\lambda$ controlling the extent of sparsity -- with higher $\lambda$ implying more sparsity, i.e. fewer features from the representation vector are selected for the final classifier. 
It is to be noted that this way of training the LR model on pre-trained RNN features is equivalent to freezing the parameters of all the hidden layers of the pre-trained RNN while tuning the parameters of a new final classification layer. 
The sparsity constraint ensures that only a small number of parameters are to be tuned which is useful to avoid overfitting when labeled data is small.

\section{Experimental Evaluation}\label{sec:exp}
We evaluate the proposed approach on binary classification tasks as defined in \cite{harutyunyan2017multitask}: i) estimating the presence (class 1) or absence (class 0) of a phenotype (e.g. cardiac dysrhythmia, chronic kidney disease, etc.) from time series of parameters such as heart rate and respiratory rate, and ii) in-hospital mortality prediction where the goal is to predict whether the patient will survive or not given time series observations after ICU admission (class 1: patient dies, class 0: patient survives). 
\newline\newline
\textbf{Dataset details}\newline
We use MIMIC-III (v1.4) clinical database \cite{johnson2016mimic} which consists of over 60,000 ICU stays across 40,000 critical care patients. 
We use benchmark data from \cite{harutyunyan2017multitask} with same data-splits for train, validation and test datasets\footnote{Refer \cite{harutyunyan2017multitask} and https://github.com/yerevann/mimic3-benchmarks for dataset sizes and other details.}.
Train, validation and test sets for various scenarios considered in our experiments are subsets of the respective original datasets (as described later).
The data contains multivariate time series for multiple physiological parameters with 12 real-valued (e.g. blood glucose level, systolic blood pressure, etc.) and 5 categorical parameters (e.g. Glascow coma scale motor response, Glascow coma scale verbal, etc.), sampled at 1 hour interval. 
The categorical variables are converted to one-hot vectors such that final multivariate time series has dimension $n=76$. 
We use time series from only up to first $48$ hours of ICU stay for all predictions (such that $\tau=48$) to imitate the practical scenario where early predictions are important.

The benchmark dataset contains label information for presence/absence of 25 phenotypes common in adult ICUs (e.g. acute cerebrovascular disease, diabetes mellitus with complications, gastrointestinal hemorrhage, etc.). 
We consider $K=20$ phenotypes to obtain the pre-trained RNN which we refer to as MIMIC-Net (MN), and test the transferability of the features from MN to remaining 5 phenotype (binary) classification tasks with varying labeled data sizes.
Since more than one phenotypes may be present in a patient at a time, we remove all patients with any of the 5 test phenotypes from the original train and validate sets (despite of them having one of the 20 train phenotypes also) to avoid any information leakage.
We report average results in terms of weighted AUROC (as in \cite{harutyunyan2017multitask}) on two random splits of 20 train phenotypes and 5 test phenotypes, such that we have 10 test phenotypes (tested one-at-a-time).
We also test transferability of MN features to in-hospital mortality prediction task.

We consider number of hidden layers $L=2$, batch size of 128, regularization using dropout factor \cite{pham2014dropout} of 0.3, and Adam optimizer \cite{kingma2014adam} with initial learning rate $10^{-4}$ for training RNNs. 
The number of hidden units $h$ with minimum $\mathcal{L}$ (eq. \ref{eq:RNN-C}) on the validation set is chosen from $\{100,200,300,400\}$. 
Best MN model was obtained for $h=300$ such that total number of features is $m=600$.
The L$_1$ parameter $\lambda$ is tuned on $\{0.1,1.0$,$\ldots$,$10^4\}$ (on a logarithmic scale) to minimize $\mathcal{L'}$ (eq. \ref{eq:RNN-C-tune}) on the validation set. 
\begin{figure}[ht]
\centering
\subfigure[\label{fig:phen}Phenotyping]{\includegraphics[width=0.5\columnwidth,trim={0cm 0cm 0cm 0cm},clip]{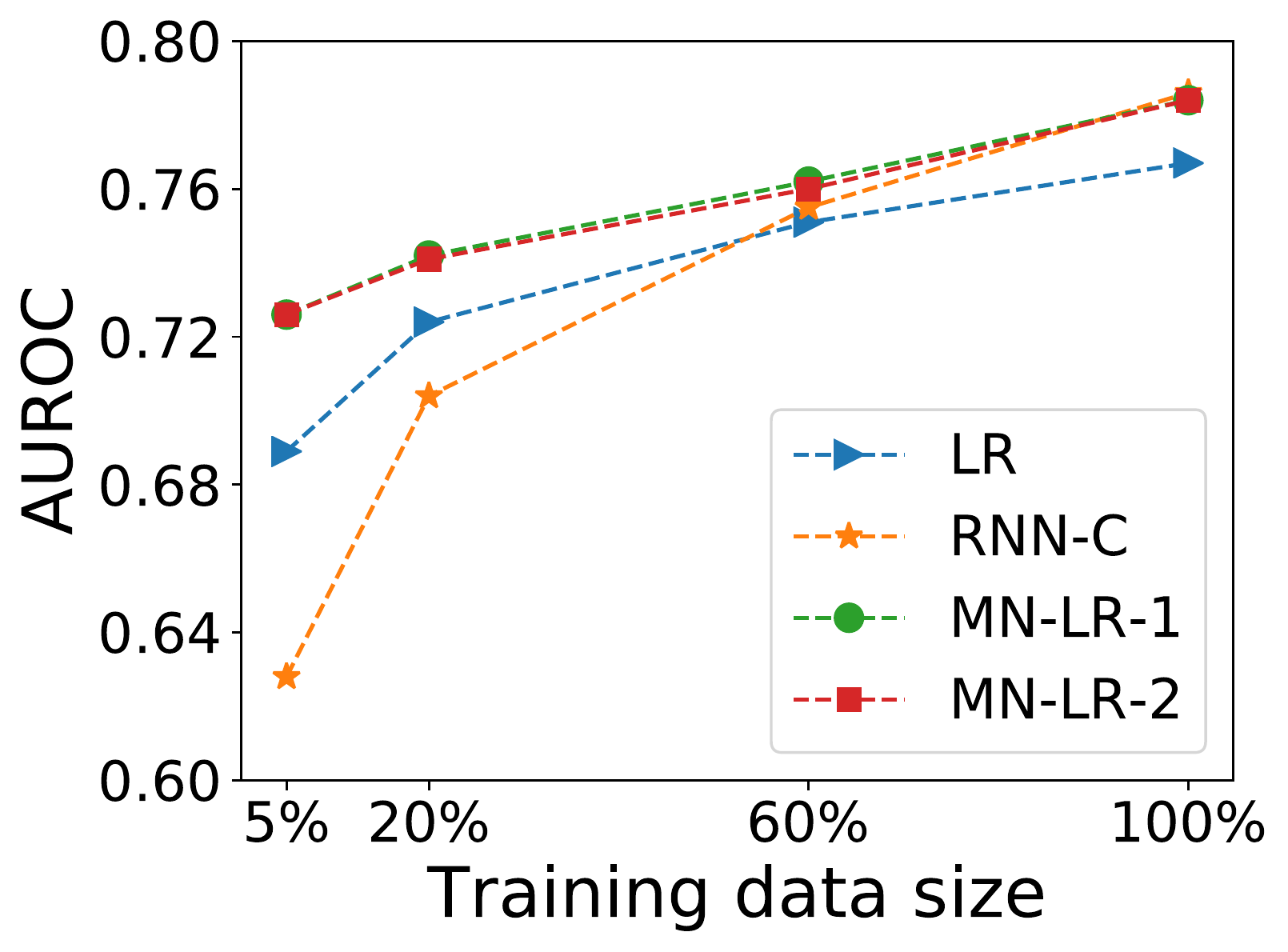}}~~
\subfigure[\label{fig:mor}In-hospital mortality prediction]{\includegraphics[width=0.5\columnwidth,trim={0cm 0cm 0cm 0cm},clip]{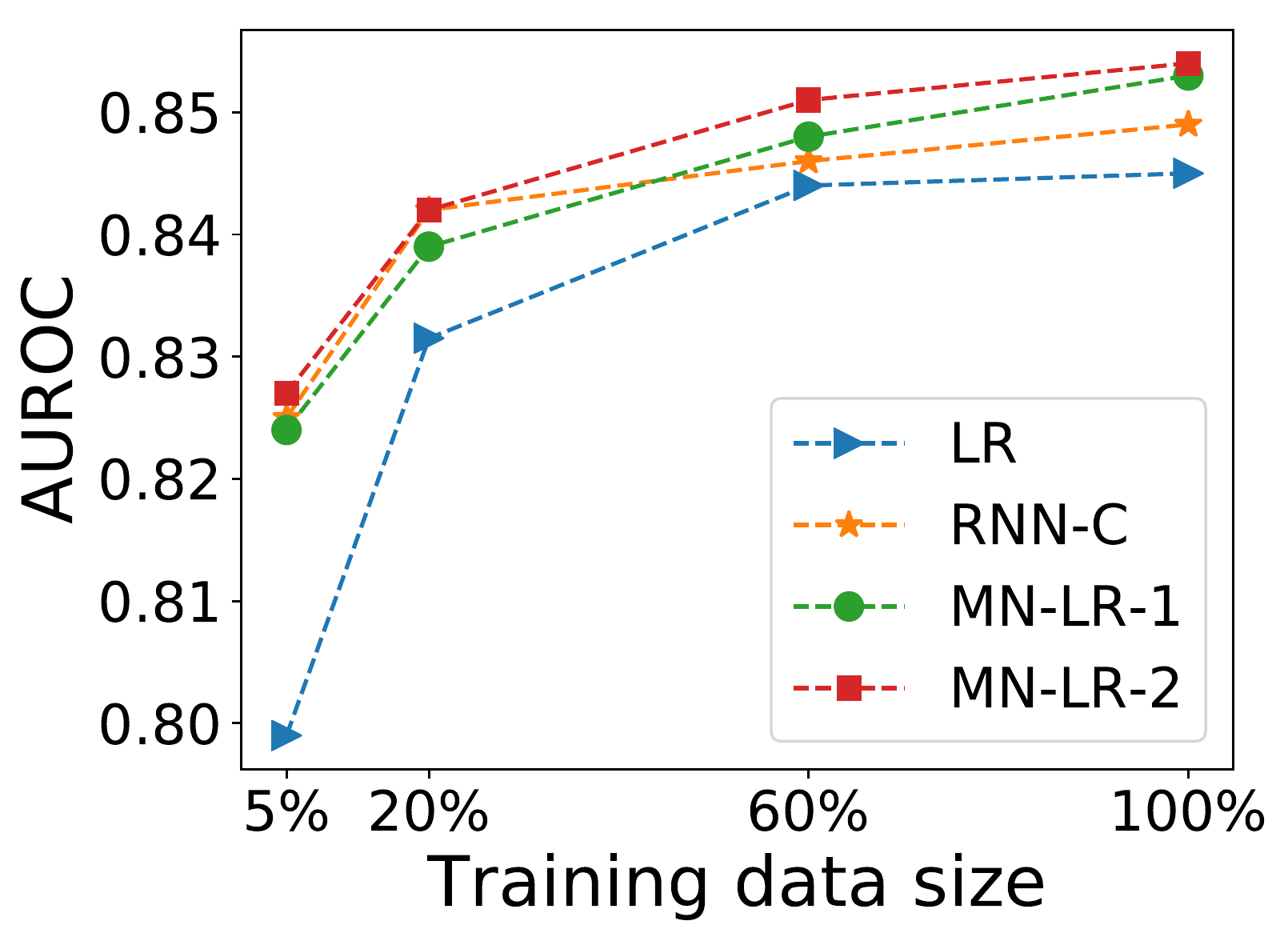}
}
\caption{AUROC with varying labeled data size. Models using transfer learning (MN-LR-1 and MN-LR-2) perform significantly better for smaller training data sizes.}
\end{figure}
\begin{table}[h]
\footnotesize
\centering
\footnotesize
\caption{Fraction of features with weight $\approx$ 0. \label{tab:sparsity}}
\begin{tabular}{|c|c|c|c|}
 \hline
{\bfseries Task}&{\bfseries LR}&{\bfseries MN-LR-1}&{\bfseries MN-LR-2}\\
\hline
Phenotyping\footnote{The average and standard deviation over 10 phenotypes is reported. }&0.902 $\pm$ 0.023&0.955 $\pm$ 0.020&0.974 $\pm$ 0.011 \\
\hline
In-hospital mortality&0.917&0.787&0.867\\
\hline
\end{tabular}
\end{table}
\newline\newline
\textbf{Results and Observations}\newline
We refer to the LR model learned using MN features as \textbf{MN-LR}, and consider two baselines for comparison: 1) Logistic Regression (\textbf{LR}) using statistical features (including mean, standard deviation, etc.) from raw time series as used in \cite{harutyunyan2017multitask}, 2) RNN classifier (\textbf{RNN-C}) learned using training data for the target task.
To test the robustness of the models for small labeled training sets, we consider subsets of training and validation datasets, while the test set remains the same.
Further, we also evaluate the relevance of layer-wise features $\mathbf{z}_{\tau,l}$ from the $L=2$ hidden layers.
\textbf{MN-LR-1} and \textbf{MN-LR-2} refer to models trained using $\mathbf{z}_{\tau,2}$ (the topmost hidden layer only) and $\mathbf{z}_{\tau} = [\mathbf{z}_{\tau,1},\mathbf{z}_{\tau,2}]$ (from both hidden layers), respectively.

\textbf{Robustness to training data size}: 
Phenotyping results in Figure \ref{fig:phen} suggest that: (i) MN-LR and RNN-C perform equally well when using 100\% training data, and are better than LR.
This implies that \textit{the transfer learning based models are as effective as models trained specifically for the target task on large labeled datasets}. (ii)
MN-LR consistently outperforms RNN-C and LR models as training dataset is reduced.
As the size of labeled training set reduces, the performance of RNN-C as well as MN-LR degrades. However, importantly, we observe that MN-LR degrades more gracefully and performs better than RNN-C. 
\textit{The performance gains from transfer learning are greater when the training set of the target task is small.
Therefore, with transfer learning, fewer labeled instances are needed to achieve the same level of performance as model trained on target data alone.}
(iii) As labeled training set is reduced, LR performs better than RNN-C confirming that deep networks are prone to overfitting on small datasets.

From Figure \ref{fig:mor}, we interestingly observe that MN-LR results are at least as good as RNN-C and LR on the seemingly unrelated task of mortality prediction, suggesting that \textit{the features learned are generic enough and transfer well.}
\begin{figure*}[th]
\centering
\includegraphics[trim={0cm 0.5cm 0cm 0cm},clip,width=\textwidth]{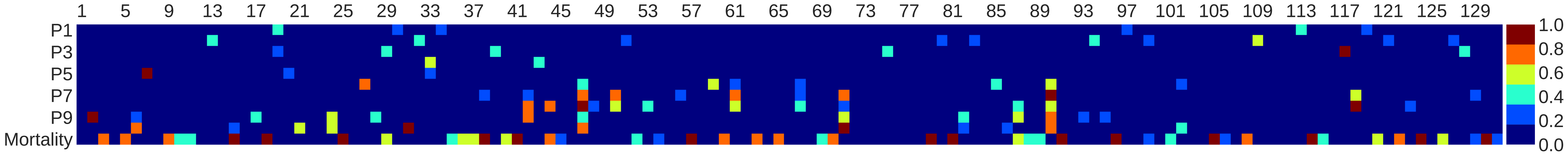}
\caption{\label{fig:sparsity} Feature weights (absolute) for MN-LR-1. Here P$_i$ ($i=1,\ldots,10$) denotes $i$-th phenotype identification task. x-axis: Feature Number, y-axis: Task.}
\end{figure*}

\textbf{Importance of features from different hidden layers}: We observe that MN-LR-1 and MN-LR-2 perform equally well for phenotyping task (Figure \ref{fig:phen}), suggesting that adding features $\mathbf{z}_{\tau,1}$ from lower hidden layer do not improve the performance given higher layer features $\mathbf{z}_{\tau,2}$. 
For the mortality prediction task, we observe slight improvement in MN-LR-2 over MN-LR-1, i.e. adding lower layer features helps. 
A possible explanation for this behavior is as follows: since training was done on phenotyping tasks, features from top-most layer suffice for new phenotypes as well; on the other hand, the more generic features from the lower layer are useful for the unrelated task of mortality prediction. 

\textbf{Number of relevant features for a task}: We observe that only a small number of features are actually relevant for a target classification task out of large number of input features to LR models (714 for LR, 300 for MN-LR-1, and 600 for MN-LR-2), 
As shown in Table \ref{tab:sparsity}, $>$95\% of features have weight $\approx$ 0 (absolute value $<$ 0.001) for MN-LR models corresponding to phenotyping tasks due to sparsity constraint (eq. \ref{eq:RNN-C-tune}), i.e. most features do not contribute to the classification decision.
The weights of features that are non-zero for at least one of target tasks for MN-LR-1 are shown in Figure \ref{fig:sparsity}.
We observe that, for example, for MN-LR-1 model only 130 features (out of 300) are relevant across the 10 phenotype classification tasks and the mortality prediction task. 
\textit{This suggests that MN provides several generic features while LR learns to select the most relevant ones given a small labeled dataset.}
Table \ref{tab:sparsity} and Figure \ref{fig:sparsity} also suggest that MN-LR models use larger number of features for mortality prediction task, possibly because concise features for mortality prediction are not available in the learned set of features as MN was pre-trained for phenotype identification tasks.

\section{Conclusion}\label{sec:future}
We have proposed an approach to leverage deep RNNs for small labeled datasets via transfer learning. 
We trained an RNN model to identify several phenotypes via multi-label classification. 
This model is found to be generalize well for new tasks including identification of new phenotypes, and interestingly, for mortality prediction.
We found that transfer learning performs better than the models trained specifically for the end task. 
Such transfer learning approaches can be a good starting point when building models with limited labeled datasets.
Transferability and generalization capability of RNNs trained simultaneously on diverse tasks (such as length of stay, mortality prediction, phenotyping, etc. \cite{harutyunyan2017multitask,song2017attend}) to new tasks is an interesting future direction.

\bibliographystyle{ACM-Reference-Format}
\bibliography{BibTex/phm-kdd2017,BibTex/nips2016,BibTex/online-ad,BibTex/ijcai2017,BibTex/kdd2018,BibTex/isense,BibTex/kdhd18,BibTex/mlmh18}

\end{document}